# A Computational Map of the Disaster Mental Health Research Landscape

Soheil Shapouri[1], Saber Soleymani[2]

[1]Department of Psychology, Lehigh University

[2]Robinson College of Business, Georgia State University

## Author Note

Soheil Shapouri https://orcid.org/0000-0001-9561-9751

We have no conflict of interest to report.

This research was supported by a Faculty Research Fund from the Department of Psychology at Lehigh University to Soheil Shapouri.

Correspondence concerning this article should be addressed to Soheil Shapouri, Fielding School of Public Health, UCLA, 650 Charles E Young Dr S, Los Angeles, CA 90095.

soheilshapouri@ucla.edu

**Abstract**

When disasters strike, they not only damage infrastructure, buildings, and the economy, but also affect the mental health of affected populations. Research on the mental health effects of disasters has expanded rapidly and has become an important, well-funded area of psychology, public health, and related social sciences. However, the overall structure of this research landscape has not been systematically characterized. Using large language models, we conducted a review of all articles related to terms “disaster” and “mental health” indexed in PubMed. 7,859 articles’ metadata were retrieved using the NCBI API, and then Claude Sonnet was used to extract structured variables from the abstracts. The results indicated that earthquakes and epidemics were the most commonly studied disasters while many other disasters such as heat and cold waves have received less attention. PTSD, depression, and anxiety were well investigated but many other mental health issues (e.g., aggression, suicide) are relatively ignored. Despite enjoying relatively large sample sizes (median = 452, IQR = 142 – 1,568), the field rarely utilizes machine learning algorithms. Hypothesis significance testing, reviews, qualitative methods, and structural equation modeling are respectively most common methods while agent-based modeling, machine learning, casual inference, and NLP each appear in less than 1% of articles. Studies are limited to a small number of countries, particularly the United States, Japan, and China, while many disaster-prone regions remain substantially underrepresented. This LLM-assisted computational map provides a broad overview of the disaster mental health literature while highlighting important methodological gaps, with particular emphasis on the underutilization of computational methods that could help advance the field.

## A Computational Map of the Disaster Mental Health Research Landscape

The frequency and intensity of several types of disasters such as wildfires, floods, and extreme weather conditions, as well as their associated economic cost has been on the rise and are projected to increase in near future (Cunningham et al., 2024; Saueret al., 2024; Newman & Noy, 2023). But disasters are not just physical or economic events, they also have societal cost and cause severe mental health problems. The mental health effects of disasters such as PTSD, depression, and anxiety begin immediately after disasters (Heanoy & Brown, 2024), affect more than 22% of the exposed population, and can last over a decade (Dückers et al., 2026). Considering large-scale catastrophic consequences of disasters, an enormous body of literature has been devoted to understanding the relationship between disasters and mental health.

Despite the substantial growth of disaster mental health research, important gaps remain in our understanding of the field: What is the current state of the field? Which topics have received the greatest attention, which populations remain understudied, and what methodological patterns characterize the existing literature? Traditionally, questions like these have been addressed through literature reviews and bibliometric analyses.

Literature review is the synthesis and evaluation of existing knowledge and published papers relevant to a particular topic. The most important form of literature review, systematic review, emerged in the 1970s, gained widespread popularity during the 1990s, and have since become a cornerstone of evidence synthesis across many scientific disciplines (Moher et al., 2015) including public health and psychology. Because they follow a predefined and transparent search strategy, systematic reviews reduce selection bias in identifying relevant studies (Uman, 2011), and when accompanied by meta-analysis, they can quantitatively synthesize evidence and estimate the overall magnitude of effects across studies. When conducted rigorously, for example by following the PRISMA guidelines (Moher et al., 2009), systematic reviews can provide valuable insights by identifying gaps in the literature, synthesizing existing evidence, and helping researchers stay up to date with developments in their field. However systematic reviews are typically focused on a narrow topic. They might be restricted to specific disasters such as hurricanes and floods (e.g., Miller et al., 2025), focused on a specific event like the 2010 earthquake in Haiti (e.g., Cénat et al., 2020), or limited to specific populations such as medical workers (Tahernejad et al., 2023). Their typically narrow focus limits literature reviews' ability to characterize an entire research field and identify broader patterns that emerge across thousands of published studies over decades.

Another approach to obtaining a broad overview of a research field is bibliometric analysis. Bibliometrics is the quantitative analysis of scientific publications using statistical methods to characterize the structure, development, and impact of a research field (Lyu et al., 2026). These analyses are typically conducted by quantitatively analyzing publication metadata, such as authors, citations, journals, institutions, and keywords, to identify patterns and trends within a research field. Bibliometrics can help researchers to find emerging trends, identify patterns of collaborations, and evaluate performance of scientists, journals, and institutions (Donthu et al., 2021).

Bibliometric analyses of disaster mental health research can identify, for example, the most common publication types (e.g., Espinosa-Pinos et al., 2023), patterns of national and international collaboration, such as the high level of national but low level of international collaboration following the 2023 Syria–Turkey earthquake (Süer et al., 2025), and disparities in research output across countries, with authors from high-income countries contributing more publications on COVID-19 and mental health than those from lower-income countries (Maalouf et al., 2021). One of the most comprehensive studies in this area is *A Bibliometric Analysis of Health-Related Literature on Natural Disasters from 1900 to 2017* (Sweileh, 2019), which analyzed more than 9,000 publications. The study showed that earthquake was the most frequently occurring keyword, that when research output was standardized by GDP per capita, India, China, and the United States ranked highest in research production, and that the average number of citations per publication was 10.8. It is important to note, however, that this study examined health-related disaster research broadly rather than focusing exclusively on disaster mental health.

Despite their important contribution, bibliometric analyses have a serious limitation. They typically rely heavily on meta-data of publications such as number of authors, citation count, co-occurrence of keywords, impact factor of journals, and rarely analyze the content of the papers. When article content is analyzed, it is most often through topic modeling or thematic analysis rather than the systematic extraction of detailed study characteristics.

Recent advances in large language models (LLMs) make it possible to combine the strengths of both approaches. Similar to how systematic reviews are conducted, LLMs can analyze the substantive content of scientific articles rather than relying solely on publication metadata. At the same time, their ability to process thousands of articles enables analyses at a scale that was previously practical only for bibliometric studies. This combination makes it possible to construct a comprehensive computational map of an entire research field revealing what is being studied, how it is being studied, and which important topics and methodological approaches remain underrepresented.

Researchers have recently begun using LLMs to accelerate and automate scientific research (Scherbakov et al., 2025). For example, LLMs have been fine-tuned to extract structured knowledge from scientific papers (Dagdelen et al., 2024), used to automate systematic reviews (Schmidt et al., 2024), and employed to generate scholarly summaries (Shamsabadi et al., 2024). Here, we use LLMs to extract structured variables from abstracts of disaster mental health studies to quantify what has been studied, identify what remains underexplored, and characterize the overall structure of the field.

## Method

**Data source.** In May 2026, all articles indexed in PubMed matching the search terms “disaster” AND “mental health” were retrieved using the NCBI E-utilities (NCBI API). A total of 8,586 records, including publication metadata, titles, and abstracts, were retrieved. Because the subsequent analyses required abstract text, 727 records without an abstract were excluded, resulting in a final dataset of 7,859 articles.

**LLM-assisted variable extraction.** The title and abstract of each article were submitted to Claude Sonnet 4.5 through the Anthropic API with the temperature set to 0 to ensure deterministic responses. A detailed prompt containing explicit coding rules, operational definitions, and predefined response categories was developed through an iterative process of prompt refinement and testing. Variables that could not be extracted reliably from abstracts during pilot testing (e.g., whether disasters were characterized as human-made or natural) were excluded from the final extraction schema. For each article, the model returned a structured JSON object containing the extracted variables, including article type, article relevance, disaster type, mental health outcomes, participants' country, sample size, and primary analytical method.

**Quality assessment.** To assess the accuracy of the LLM-based extraction, 15 articles were randomly selected from the dataset and independently coded by the author using the same variable definitions and coding categories provided to the LLM. The human-coded values were then compared with the LLM-extracted values for each variable.

Across the 15 randomly selected articles, the LLM classifications showed 95.8% exact agreement with human coding across all applicable extracted variables. This represents a conservative estimate of agreement, as variables allowing multiple classifications were considered concordant only when the complete set of classifications matched exactly; any partial match, including the omission or addition of a single category, was coded as a disagreement.

## Results

Of 7,859 articles initially identified, 9 duplicate titles, 3 duplicate DOIs, and 6 records with parsing errors were removed; after restricting the dataset to disaster and mental health research, 5,614 articles remained for analysis, spanning more than five decades of research from 1970 to 2026. Most articles were empirical quantitative studies (55.9%), followed by reviews (12.0%), policy/commentary/editorial articles (10.0%), and qualitative studies (6.7%).

The literature was dominated by studies of natural disasters with earthquakes (22.35%), epidemics/pandemics (17.76%), hurricanes/storms (12.13%), floods (8.23%), and wildfires (4.67%) being the five most frequently investigated specific disaster types while other natural disasters such as extreme temperatures (1.82%) have not received same level of attention. The most frequently studied technological disasters were radiation events (3.1%) and transport accidents (1.6%). Overall, 472 articles (8.4%) examined multiple disaster types, with substantial heterogeneity in the combinations of disasters studied (hurricanes and oil spill, epidemics and earthquakes, etc.) and no single combination standing out.

The number of disaster mental health publications increased substantially after 2000 (Figure X, Panel A), a pattern observed across most major disaster types (Panel B). Epidemic/pandemic research showed a distinct trajectory, increasing sharply after 2019 and subsequently declining toward pre-pandemic levels.

Figure 1

*Temporal Trends in Disaster Mental Health Research*

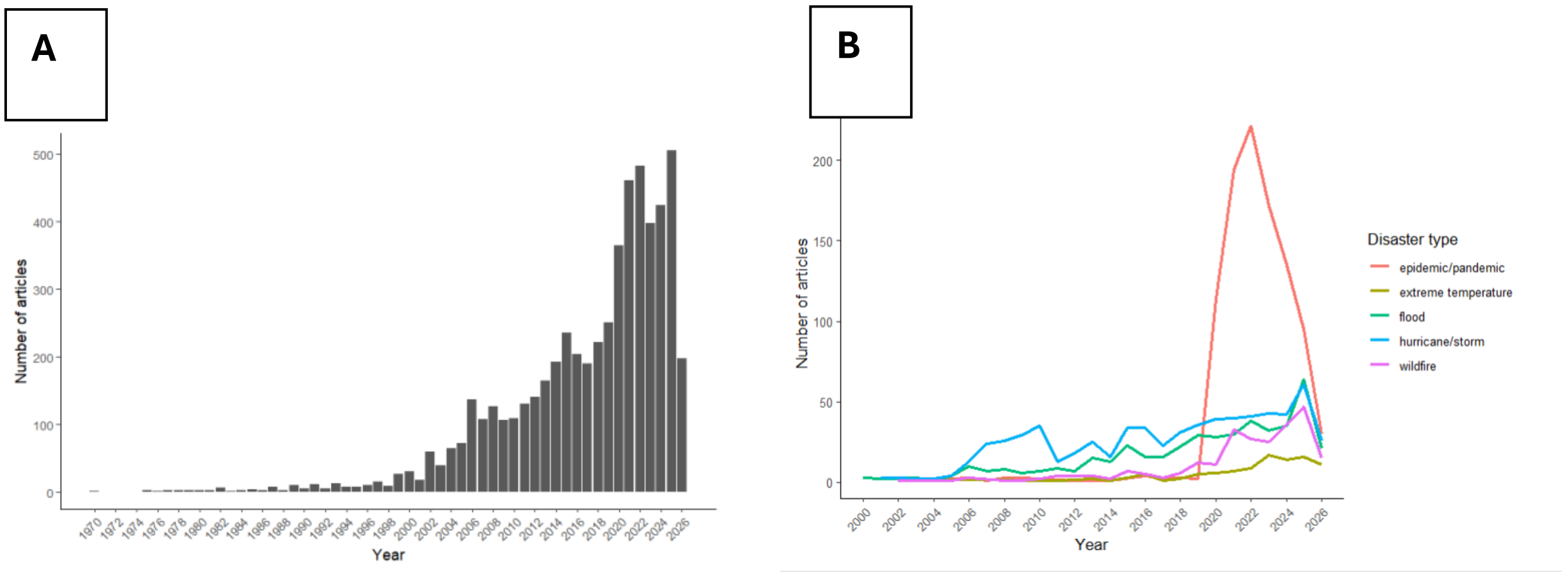

Figure 1. Trends in disaster mental health research over time. (A) Annual number of disaster mental health articles published between 1970 and 2026. (B) Annual number of articles examining selected major disaster types between 2000 and 2026. Data for 2026 are incomplete because records were collected in May 2026.

A considerable proportion of studies (35.3%) examined multiple mental health outcomes. Across all outcomes examined, PTSD was the most common followed by depression and anxiety, whereas aggression, suicidality, and substance use received considerably less attention (Table 1).

**Table 1**

*Mental Health Outcomes Examined Across Disaster Research*

| Mental health outcome | n | % |
|---|---|---|
| PTSD | 2,082 | 37.1 |
| Depression | 1,713 | 30.5 |
| Anxiety/stress | 1,390 | 24.8 |
| Distress | 1,331 | 23.7 |
| Resilience/growth | 476 | 8.5 |
| Substance use | 267 | 4.8 |
| Suicidality | 266 | 4.7 |
| Aggression | 38 | 0.7 |

*Note.* Articles could examine more than one mental health outcome; therefore, percentages are not mutually exclusive and do not sum to 100%.

Across all 5,614 articles, the participant country was not applicable for 1,121 (19.97%) and unclear for 476 (8.48%). Among identified countries, the United States was most common (1,038; 18.49%), followed by Japan (461; 8.21%) and China (420; 7.48%). An additional 316 articles (5.63%) involved multiple countries, while all other individual countries each accounted for less than 4% of the articles.

When all reported countries were considered, the same pattern emerged, with the United States most frequently represented (19.2%), followed by Japan (8.5%), China (7.7%), Australia (3.8%), and Türkiye (3.1%); all other individual countries accounted for less than 2% of articles.

Sample sizes were available for 3,362 articles (59.9%) and articles with sample sizes classified as not applicable (n = 1,479; 26.3%) or unclear (n = 772; 13.8%) were excluded. Among articles which reported sample size, the median was 452 (IQR = 142–1,568). Sample sizes ranged from as few as one participant in case studies to several million individuals in large national cohorts, with exceptionally large values reflecting weighted population estimates or analyses of large-scale data such as millions of social media posts.

Hypothesis testing/classical statistics was the predominant analytical approach, appearing in 3,002 articles (53.5%), followed by reviews (12.2%), qualitative methods (9.5%), and structural equation modeling/path analysis (5.1%). Other analytical approaches were rarely used, including machine learning (0.84%), causal inference (0.78%), NLP (0.43%), Bayesian statistics (0.12%), neural networks/deep learning (0.07%), and agent-based modeling (0.04%).

## Discussion

Our study provides a comprehensive overview of disaster mental health research which has had a notable growth since 2000. Inspired by studies that have shown high accuracy and efficiency of LLMs in data extraction from scientific papers (Gartlehner et al., 2025b) we used Claude Sonnet 4.5 to generate a quantitative picture of the disaster and mental health research landscape. The data extraction part took only a few hours, required minimal human effort, and cost under $200, showing that this approach may offer an efficient and accessible way to map other research fields and identify their major trends and gaps.

Our findings reveal that natural hazards have captured more attention than technological catastrophes with earthquakes and epidemics being the most frequent disasters that are studied. Considering the possibility of more severe mental health issues caused by technological disasters than natural disasters (Bromet et al., 2017) the field should pay greater attention to the former.

PTSD, depression, and anxiety remain the hottest topic of the field, a pattern that might reflect inclusion of these disorders in Diagnostic and Statistical Manual of Mental Disorders (DSM), availability of validated measures such as PCL, PHQ, and GAD to easily assess these disorders with a handful of questions (Moore et al., 2020), or simply continuity of established lines of research. Relative ignorance of other mental health outcomes such as aggression, suicide, and substance use are also evident in our results. Researchers have already developed theoretical frameworks, such as social ecological framework to better understand the violence in disasters contexts (Medzhitova et al., 2023), that should be utilized more often to move the field forward toward new insights.

Our analysis indicates that the average sample size in the field is larger than typical psychology fields. The average sample size is 104 in social-personality psychology (Fraley & Vazire, 2014), and 179 in clinical psychology research (Reardon et al., 2019). While psychological research sample sizes are gradually increasing (Fraley et al., 2022) disaster mental health research has already enjoyed relatively larger samples. However, this strength of the field has not been fully leveraged through the adoption of advanced analytical methods, such as machine learning.

With the exception of structural equation modeling and path analysis, the field is methodologically dominated by hypothesis significance testing, qualitative studies, and reviews, leaving more advanced computational techniques underutilized. This is especially surprising for two reasons: First, many studies have large enough sample sizes typically required for other methods such as machine learning, network analysis, and Bayesian hierarchical modeling. Second, public health is an applied field where prediction of outcomes might be much more useful than inference about individual variables.

Although our findings are based on an advanced analytical approach and thousands of published papers, limitations regarding the database used for this study warrant caution when interpreting the observed patterns. We used PubMed which provides free, fast API for downloading meta-data of papers, but it indexes biomedical and life sciences literature. Considering the variation of coverage by different bibliographical databases (Gusenbauer, 2022), some of the patterns reported here might have differed had we used another database. For example, IEEE Xplore has a few disaster mental health papers not indexed by PubMed. However, it is important to note that IEEE Xplore is not commonly used by social scientists, and they better reflect computer science and engineering than public health and psychology. PsycInfo could be another database used for this research but while PubMed and PsycInfo coverage are sometimes different (Löhönen et al., 2010), PsycInfo does not allow unrestricted, free access provided by PubMed.

The efficiency of LLMs in data extraction (Gartlehner et al., 2025a) has enabled researchers to obtain a broad overview of an entire field quickly. We used this approach to provide a quantitative summary of the disaster mental health literature, identifying the most commonly studied variables and analytical methods, as well as understudied topics and underutilized techniques. This approach offers a promising avenue for systematically identifying research gaps and guiding future research priorities across scientific disciplines.